# Penta and Hexa Valued Representation of Neutrosophic Information

Vasile Pătraşcu

Tarom Information Technology, Bucharest-Otopeni, Romania.

E-mail: patrascu.v@gmail.com

**Abstract.** Starting from the primary representation of neutrosophic information, namely the degree of truth, degree of indeterminacy and degree of falsity, we define a nuanced representation in a penta valued fuzzy space, described by the index of truth, index of falsity, index of ignorance, index of contradiction and index of hesitation. Also, it was constructed an associated penta valued logic and then using this logic, it was defined for the proposed penta valued structure the following operators: union, intersection, negation, complement and dual. Then, the penta valued representation is extended to a hexa valued one, adding the sixth component, namely the index of ambiguity.

## 1 Introduction

The neutrosophic representation of information was proposed by Florentin Smarandache [6], [13-19] and it is a generalisation of intuitionistic fuzzy representation proposed by Krassimir Atanassov [1-4] and also for fuzzy representation proposed by Lotfi Zadeh [20]. The neutrosophic representation is described by three parameters: degree of truth μ, degree of indeterminacy ω and degree of falsity ν. In this paper we assume that the parameters $\mu, \omega, \nu \in [0,1]$. The representation space $(\mu, \omega, \nu)$ is a primary space for neutrosophic information. Starting from primary space, it can be derived other more nuanced representations belonging to multi-valued fuzzy spaces where the set of parameters defines fuzzy partitions of unity. In these multi-valued fuzzy spaces, at most four parameters of representation are different from zero while all the others are zero [7], [8], [9], [10]. In the following, the paper has the structure: Section 2 presents the construction of two multi-valued representation for bipolar information. The first is based on Belnap logical values, namely true, false, unknown and contradictory while the second is based on a new logic that was obtained by adding to the Belnap logic the fifth value: ambiguity; Section 3 presents two variants for penta valued representation of neutrosophic information based on truth, falsity, ignorance, contradiction and hesitancy; Section 4 presents a penta valued logic that uses the following values: true, false, unknown, contradictory and hesitant; Section 5 presents five operators for the penta valued structures constructed in section 3. Firstly, it was defined two binary operators namely union and intersection, and secondly, three unary operators, namely complement, negation and dual. All these five operators where defined in concordance with the logic presented in the section 4; Section 6 extend the penta valued structures presented in section 3 by filtering from truth and falsity the sixth feature, namely the ambiguity; The last section outlines some conclusions.

## 2    Tetra and Penta Valued Representation of Bipolar Information

The bipolar information is defined by the degree of truth $\mu$ and the degree of falsity $\nu$. Also, it is associated with a degree of certainty and a degree of uncertainty. The bipolar uncertainty can have three features well outlined: ambiguity, ignorance and contradiction. All these three features have implicit values that can be calculated using the bipolar pair $(\mu, \nu)$. In the same time, ambiguity, ignorance and contradiction can be considered features belonging to indeterminacy but to an implicit indeterminacy. We can compute the values of these implicit features of indeterminacy. First we calculate the index of ignorance $\pi$ and index of contradiction $\kappa$:

$$\pi = 1 - \min(1, \mu + \nu) \qquad (2.1)$$

$$\kappa = \max(1, \mu + \nu) - 1 \qquad (2.2)$$

There is the following equality:

$$\mu + \nu + \pi - \kappa = 1 \qquad (2.3)$$

which turns into the next tetra valued partition of unity:

$$(\mu - \kappa) + (\nu - \kappa) + \pi + \kappa = 1 \qquad (2.4)$$

The four terms form (2.4) are related to the four logical values of Belnap logic: true, false, unknown and contradictory [5]. Further, we extract from the first two terms the bipolar ambiguity $\alpha$:

$$\alpha = 2 \cdot \min(\mu - \kappa, \nu - \kappa) \qquad (2.5)$$

Moreover, on this way, we get the two components of bipolar certainty: index of truth $\tau^+$ and index of falsity $\tau^-$:

$$\tau^+ = \mu - \kappa - \frac{\alpha}{2} \qquad (2.6)$$

$$\tau^- = \nu - \kappa - \frac{\alpha}{2} \qquad (2.7)$$

So, we obtained a penta valued representation of bipolar information by $(\tau^+, \tau^-, \alpha, \pi, \kappa)$. The vector components verify the partition of unity condition, namely:

$$\tau^+ + \tau^- + \alpha + \pi + \kappa = 1 \qquad (2.8)$$

Formula (2.8) suggests the bipolar information structure shown in figure 1.

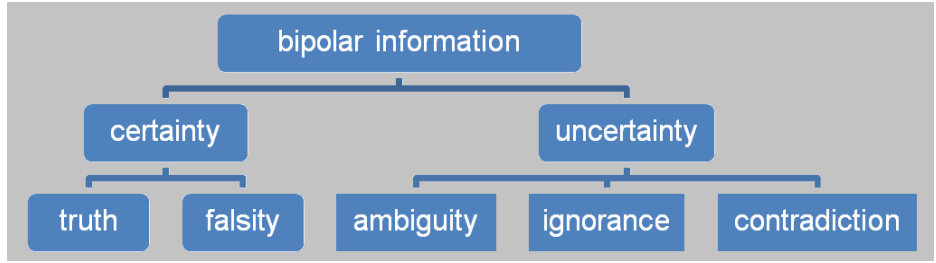

**Fig. 1.** The bipolar information structure.

In the following sections, the two representations defined by (2.4) and (2.8) will be used to represent the neutrosophic information in two penta valued structures.

## 3   Penta Valued Representation of Neutrosophic Information Based on Truth, Falsity, Ignorance, Contradiction and Hesitation

In this section we present two variants for this type of penta valued representation of neutrosophic information.

### 3.1   Variant (I)

Using the penta valued partition (2.8), described in Section 3, first, we construct a partition with ten terms for neutrosophic information and then a penta valued one, thus:

$$(\tau^+ + \tau^- + \alpha + \pi + \kappa)(\omega + 1 - \omega) = 1 \qquad (3.1.1)$$

By multiplying we obtain ten terms that describe the following ten logical values: weak true, weak false, neutral, saturated, hesitant, true, false, unknown, contradictory and ambiguous: $t_w = \omega\tau^+, f_w = \omega\tau^-, n = \omega\pi, s = \omega\kappa, h = \omega\alpha,$

$t = (1-\omega)\tau^+, f = (1-\omega)\tau^-, u = (1-\omega)\pi, c = (1-\omega)\kappa, a = (1-\omega)\alpha.$

The first five terms refer to the upper square of the neutrosophic cube while the next five refer to the bottom square of the neutrosophic cube (fig. 2). We distribute equally the first four terms between the fifth and the next four and then the tenth, namely the ambiguity, equally, between true and false and we obtain:

$$t = (1-\omega)\tau^+ + \frac{\omega\tau^+}{2} + \frac{(1-\omega)\alpha}{2}$$

$$f = (1-\omega)\tau^- + \frac{\omega\tau^-}{2} + \frac{(1-\omega)\alpha}{2}$$

$$u = (1-\omega)\pi + \frac{\omega\pi}{2}$$

$$c = (1-\omega)\kappa + \frac{\omega\kappa}{2}$$

$$h = \omega\alpha + \frac{\omega\tau^+}{2} + \frac{\omega\tau^-}{2} + \frac{\omega\pi}{2} + \frac{\omega\kappa}{2}$$

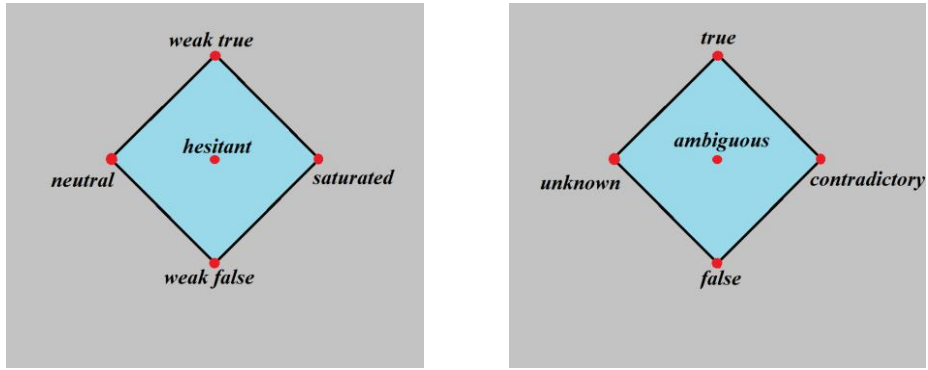

**Fig. 2.** The upper and the bottom square of neutrosophic cube and there logical values.

then, we get the following equivalent form for the five final parameters:

$$t = \left(1 - \frac{\omega}{2}\right)(\mu - \kappa) - \frac{\omega\alpha}{4} \quad (3.1.2)$$

$$f = \left(1 - \frac{\omega}{2}\right)(\nu - \kappa) - \frac{\omega\alpha}{4} \quad (3.1.3)$$

$$u = \left(1 - \frac{\omega}{2}\right)\pi \quad (3.1.4)$$

$$c = \left(1 - \frac{\omega}{2}\right)\kappa \quad (3.1.5)$$

$$h = \frac{(1+\alpha)}{2}\omega \quad (3.1.6)$$

The five parameters defined by relations (3.1.2-3.1.6) define a partition of unity:

$$t + f + h + c + u = 1 \quad (3.1.7)$$

Thus, we obtained a penta valued representation of neutrosophic information based on logical values: true, false, unknown, contradictory and hesitant. Since $\pi \cdot \kappa = 0$, it results that $u \cdot c = 0$ and hence the conclusion that only four of the five terms from the partition can be distinguished from zero. Geometric representation of this construction can be seen in figure 3.

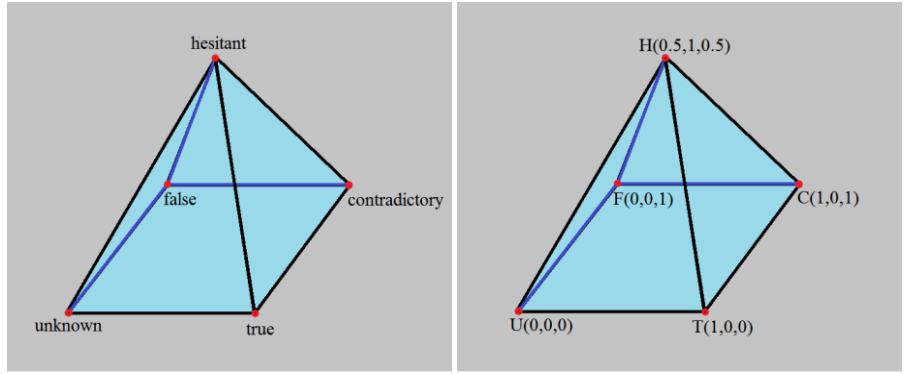

**Fig. 3.** The geometrical representation of the penta-valued space, based on true, false, unknown, contradictory and hesitant.

**The Inverse Transform.** There exist a method to compute the inverse transform from penta-valued representation $(t, f, h, c, u)$ to the primary representation $(\mu, \omega, \nu)$. This method will not be subject of this paper. It results the following formulas:

$$\mu = \frac{1}{2} + \frac{t - f + c - u}{2 - \beta + \sqrt{\beta^2 - 2h}} \qquad (3.1.8)$$

$$\omega = \beta - \sqrt{\beta^2 - 2h} \qquad (3.1.9)$$

$$\nu = \frac{1}{2} + \frac{f - t + c - u}{2 - \beta + \sqrt{\beta^2 - 2h}} \qquad (3.1.10)$$

where:

$$\beta = \frac{1}{2} + h + \min(t, f) \qquad (3.1.11)$$

### 3.2 Variant (II)

Using the tetra-valued partition defined by formula (2.4) we obtain:

$$\frac{\left(\mu - \kappa - \frac{\alpha\omega}{2}\right) + \left(\nu - \kappa - \frac{\alpha\omega}{2}\right) + \pi + \kappa + \omega}{1 + \omega - \alpha\omega} = 1 \qquad (3.2.1)$$

It results a penta valued partition of unity for neutrosophic information. These five terms are related to the following logical values: true, false, unknown, contradictory, hesitation:

$$t = \frac{\mu - \kappa - \frac{\alpha\omega}{2}}{1 + (1-\alpha)\omega} \qquad (3.2.2)$$

$$f = \frac{\nu - \kappa - \frac{\alpha\omega}{2}}{1 + (1-\alpha)\omega} \qquad (3.2.3)$$

$$u = \frac{\pi}{1 + (1-\alpha)\omega} \qquad (3.2.4)$$

$$c = \frac{\kappa}{1 + (1-\alpha)\omega} \qquad (3.2.5)$$

$$h = \frac{\omega}{1 + (1-\alpha)\omega} \qquad (3.2.6)$$

Formula (3.2.1) becomes:

$$t + f + h + c + u = 1 \qquad (3.2.7)$$

**The Inverse Transform.** The next three formulas represent components of the inverse transform from the penta valued space of representation to the primary one:
The values of the parameters $\mu, \omega, \nu$ are given by:

$$\mu = \frac{1}{2} + \frac{t - f + c - u}{1 + \sqrt{1 - 4h(|t-f| + |c-u|)}} \qquad (3.2.8)$$

$$\omega = \frac{2h}{1 + \sqrt{1 - 4h(|t-f| + |c-u|)}} \qquad (3.2.9)$$

$$\nu = \frac{1}{2} + \frac{f - t + c - u}{1 + \sqrt{1 - 4h(|t-f| + |c-u|)}} \qquad (3.2.10)$$

## 4   Penta Valued Logic Based on Truth, Falsity, Ignorance, Contradiction and Hesitation

This five-valued logic is a new one, but is related to our previous works presented in [11], [12]. In the framework of this logic we will consider the following five logical values: *true* $t$, *false* $f$, *unknown* $u$, *contradictory* $c$, and *hesitant* $h$. We have obtained these five logical values, adding to the four Belnap logical values the fifth: *hesitant*. Tables 1, 2, 3, 4, 5, 6 and 7 show the basic operators in this logic. The main differences between the proposed logic and the Belnap logic are related to the logical values $u$ and $c$. We have defined $c \cap u = h$ and $c \cup u = h$ while in the Belnap logic there were defined $c \cap u = f$ and $c \cup u = t$.

**Table 1.** The Union

| ∪ | t | c | h | u | f |
|---|---|---|---|---|---|
| t | t | t | t | t | t |
| c | t | c | h | h | c |
| h | t | h | h | h | h |
| u | t | h | h | u | u |
| f | t | c | h | u | f |

**Table 2.** The Intersection.

| ∩ | t | c | h | u | f |
|---|---|---|---|---|---|
| t | t | c | h | u | f |
| c | c | c | h | h | f |
| h | h | h | h | h | f |
| u | u | h | h | u | f |
| f | f | f | f | f | f |

**Table 3.** The Complement.

|   | ¬ |
|---|---|
| t | f |
| c | c |
| h | h |
| u | u |
| f | t |

**Table 4.** The Negation.

|   | — |
|---|---|
| t | f |
| c | u |
| h | h |
| u | c |
| f | t |

**Table 5.** The Dual.

|   | ≈ |
|---|---|
| t | t |
| c | u |
| h | h |
| u | c |
| f | f |

The complement, the negation and the dual are interrelated and there exists the following equalities: $\approx x = -\neg x$, $\neg x = -\approx x$, $-x = \neg \approx x$

**Table 6.** The Equivalence

| ↔ | t | c | h | u | f |
|---|---|---|---|---|---|
| t | t | c | h | u | f |
| c | c | c | h | h | c |
| h | h | h | h | h | h |
| u | u | h | h | u | u |
| f | f | c | h | u | t |

The *equivalence* is calculated by $x \leftrightarrow y = (\neg x \cup y) \cap (x \cup \neg y)$

**Table 7.** The S-implication

| → | t | c | h | u | f |
|---|---|---|---|---|---|
| t | t | c | h | u | f |
| c | t | c | h | h | c |
| h | t | h | h | h | h |
| u | t | h | h | u | u |
| f | t | t | t | t | t |

The *S-implication* is calculated by $x \rightarrow y = \neg x \cup y$

## 5  New Operators Defined on the Penta Valued Structure

There be $x = (t, c, h, u, f) \in [0,1]^5$. For this kind of vectors, one defines the union, the intersection, the complement, the negation and the dual operators. The operators are related to those define in [12].

**The Union**: For two vectors $a, b \in [0,1]^5$, where $a = (t_a, c_a, h_a, u_a, f_a)$, $b = (t_b, c_b, h_b, u_b, f_b)$, one defines the union (disjunction) $d = a \cup b$ by the formula:

$$t_d = t_a \vee t_b$$
$$c_d = (c_a + f_a) \wedge (c_b + f_b) - f_a \wedge f_b$$
$$u_d = (u_a + f_a) \wedge (u_b + f_b) - f_a \wedge f_b \quad (5.1)$$
$$f_d = f_a \wedge f_b$$

with $\quad h_d = 1 - (t_d + c_d + u_d + f_d)$

**The Intersection**: For two vectors $a, b \in [0,1]^5$ one defines the intersection (conjunction) $c = a \cap b$ by the formula:

$$t_c = t_a \wedge t_b$$
$$c_c = (c_a + t_a) \wedge (c_b + t_b) - t_a \wedge t_b$$
$$u_c = (u_a + t_a) \wedge (u_b + t_b) - t_a \wedge t_b \quad (5.2)$$
$$f_c = f_a \vee f_b$$

with $\quad h_c = 1 - (t_c + c_c + u_c + f_c)$

In formulae (5.1) and (5.2), the symbols "$\vee$" and "$\wedge$" represent the maximum and the minimum operators, namely: $\forall x, y \in [0,1]$, $x \vee y = \max(x, y)$ and $x \wedge y = \min(x, y)$. The union "$\cup$" and intersection "$\cap$" operators preserve de properties $t + c + u + f \leq 1$ and $u \cdot c = 0$, namely:

$$t_{a \cup b} + c_{a \cup b} + u_{a \cup b} + f_{a \cup b} \leq 1$$
$$c_{a \cup b} \cdot u_{a \cup b} = 0$$
$$t_{a \cap b} + c_{a \cap b} + u_{a \cap b} + f_{a \cap b} \leq 1$$
$$c_{a \cap b} \cdot u_{a \cap b} = 0$$

***The Complement***: For $x = (t, c, h, u, f) \in [0,1]^5$ one defines the complement $x^c$ by formula:
$$x^c = (f, c, h, u, t) \quad (5.3)$$

***The Negation***: For $x = (t, c, h, u, f) \in [0,1]^5$ one defines the negation $x^n$ by formula:
$$x^n = (f, u, h, c, t) \quad (5.4)$$

***The Dual***: For $x = (t, c, h, u, f) \in [0,1]^5$ one defines the dual $x^d$ by formula:
$$x^d = (t, u, h, c, f) \quad (5.5)$$

In the set $\{0,1\}^5$ there are five vectors having the form $x = (t, c, h, u, f)$, which verify the condition $t + f + c + h + u = 1$: $T = (1,0,0,0,0)$ (*True*), $F = (0,0,0,0,1)$ (*False*), $C = (0,1,0,0,0)$ (*Contradictory*), $U = (0,0,0,1,0)$ (*Unknown*) and $H = (0,0,1,0,0)$ (*Hesitant*). Using the operators defined by (5.1), (5.2), (5.3), (5.4) and (5.5), the same truth table results as seen in Tables 1, 2, 3, 4, 5, 6 and 7. Using the complement, the negation and the dual operators defined in the penta valued space and returning in the primary three-valued space, we find the following equivalent unary operators:

$$(\mu, \omega, \nu)^c = (\nu, \omega, \mu) \quad (5.6)$$
$$(\mu, \omega, \nu)^n = (1 - \mu, \omega, 1 - \nu) \quad (5.7)$$
$$(\mu, \omega, \nu)^d = (1 - \nu, \omega, 1 - \mu) \quad (5.8)$$

## 6   Hexa Valued Representation of Neutrosophic Information

In this section we will extend the two penta valued representations presented in section 3 to hexa valued representations. We will obtain two variants.

### 6.1 Variant (I)

From the penta valued structure presented in the section 3.1 we will extract the ambiguity from index of truth and index of falsity and on this way we obtain the following formulae for index of truth, index of falsity and index of ambiguity:

$$t = \left(1 - \frac{\omega}{2}\right)\tau^+ \qquad (6.1.1)$$

$$f = \left(1 - \frac{\omega}{2}\right)\tau^- \qquad (6.1.2)$$

$$a = (1 - \omega)\alpha \qquad (6.1.3)$$

The formulae for index of ignorance, contradiction and hestitation remained unchanged, namely:

$$u = \left(1 - \frac{\omega}{2}\right)\pi \qquad (6.1.4)$$

$$c = \left(1 - \frac{\omega}{2}\right)\kappa \qquad (6.1.5)$$

$$h = \frac{(1 + \alpha)}{2}\omega \qquad (6.1.6)$$

The six parameters defined by relations (6.1.1-6.1.6) define a partition of unity:

$$t + f + a + h + c + u = 1 \qquad (6.1.7)$$

Thus, we obtained a hexa valued representation of neutrosophic information based on logical values: true, false, ambiguous, unknown, contradictory and hesitant. Since $\tau^+ \cdot \tau^- = 0$ and $\pi \cdot \kappa = 0$, it results that $t \cdot f = 0$ and $u \cdot c = 0$ and hence the conclusion that only four of the six terms from the partition can be distinguished from zero. This hexa valued representation suggests the neutrosophic information structure that can be seen in figure 4.

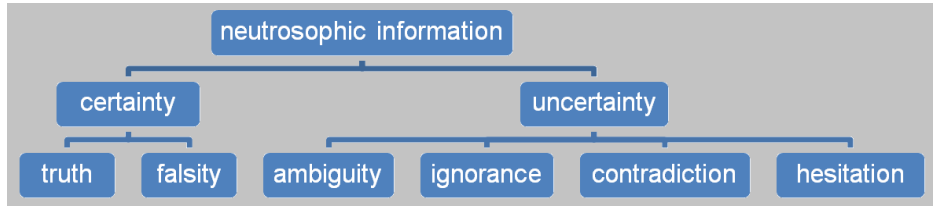

**Fig. 4.** The neutrosophic information structure.

### 6.2 Variant (II)

From the penta valued structure presented in the section 3.1 we will extract the ambiguity from index of truth and index of falsity and on this way we obtain the following formulae for index of truth, index of falsity and index of ambiguity:

$$t = \frac{\tau^+}{1 + (1-\alpha)\omega} \qquad (6.2.1)$$

$$f = \frac{\tau^-}{1 + (1-\alpha)\omega} \qquad (6.2.2)$$

$$a = \frac{(1-\omega)\alpha}{1 + (1-\alpha)\omega} \qquad (6.2.3)$$

The formulae for index of ignorance, contradiction and hestitation remained unchanged, namely:

$$u = \frac{\pi}{1 + (1-\alpha)\omega} \qquad (6.2.4)$$

$$c = \frac{\kappa}{1 + (1-\alpha)\omega} \qquad (6.2.5)$$

$$h = \frac{\omega}{1 + (1-\alpha)\omega} \qquad (6.2.6)$$

The six parameters defined by relations (6.1.1-6.1.6) define a partition of unity:

$$t + f + a + h + c + u = 1 \qquad (6.2.7)$$

Also, $t \cdot f = 0$ and $u \cdot c = 0$ and hence the conclusion that only four of the six terms from the partition can be distinguished from zero.

## 7 Conclusion

In this paper it was presented two new penta valued structures for neutrosophic information. These structures are based on Belnap logical values, namely true, false, unknown, and contradictory plus a fifth, hesitant. It defines the direct conversion from ternary space to the penta valued one and also the inverse transform from penta-valued space to the primary one. There were defined the logical operators for the penta valued structures: union, intersection, complement, dual and negation. Also the two penta valued representations was extended to hexa value representations adding the sixth logical value, namely ambiguous.